\PassOptionsToPackage{small,bf}{caption}

\documentclass[smallextended]{svjour3}
\RequirePackage{fix-cm}
\usepackage{mathptmx} 
\smartqed
\journalname{Machine Translation Journal}

\usepackage{url}
\usepackage{epsfig}
\usepackage{latexsym}
\usepackage{microtype}
\usepackage{todonotes}
\usepackage{tabularx}
\usepackage{amsmath}
\usepackage{arydshln}
\usepackage{booktabs}
\usepackage{wrapfig}
\usepackage{pgfplots}
\usepackage{textcomp}
\usepackage{chngpage}
\usepackage{graphicx,adjustbox}
\usepackage{multirow, makecell}
\usepackage[group-separator={,}, group-four-digits=true, group-digits=integer]{siunitx}
\usepackage[notext,not1]{stix}
\usepackage{fancyvrb}
\usepackage{natbib}
\usepackage{tabularx}
\usepackage{colortbl}

\usepackage[T5,T1]{fontenc}

\usepackage{tikz}
\usepackage[linguistics]{forest}

\let\citewpar\citep
\let\citewopar\citet

\titlerunning{A Survey of Orthographic Information in Machine Translation}
\authorrunning{Chakravarthi et al.}

\title{A Survey of Orthographic Information in Machine Translation }
\author{ Bharathi~Raja~Chakravarthi\(^1\) \and Priya~Rani \(^1\) \\ \and Mihael~Arcan\(^2\)  \and John~P.~McCrae \(^1\) }
\institute{
\begin{tabularx}{\textwidth}{l}
    Bharathi Raja Chakravarthi \\
    bharathi.raja@insight-centre.org\\
    \\
    Priya Rani \\
    priya.rani@insight-centre.org\\
    \\
    Mihael Arcan\\
    mihael.arcan@insight-centre.org\\
    \\
    John P. McCrae\\
    john.mccrae@insight-centre.org\\
    \\
    \(^1\) Unit for Linguistic Data, Insight Centre for Data Analytics, \\Data Science Institute, National University of Ireland Galway \\
    \(^2\) Unit for Natural Language Processing, Insight Centre for Data Analytics, \\Data Science Institute, National University of Ireland Galway  \\
\end{tabularx}
}

\begin{document}
\maketitle
\begin{abstract}
Machine translation is one of the applications of natural language processing which has been explored in different languages. Recently researchers started paying attention towards machine translation for resource-poor languages and closely related languages. A widespread and underlying problem for these machine translation systems is the variation in orthographic conventions which causes many issues to traditional approaches. Two languages written in two different orthographies are not easily comparable, but orthographic information can also be used to improve the machine translation system. This article offers a survey of research regarding orthography's influence on machine translation of under-resourced languages. It introduces under-resourced languages in terms of machine translation and how orthographic information can be utilised to improve machine translation. We describe previous work in this area, discussing what underlying assumptions were made, and showing how orthographic knowledge improves the performance of machine translation of under-resourced languages. We discuss different types of machine translation and demonstrate a recent trend that seeks to link orthographic information with well-established machine translation methods. Considerable attention is given to current efforts of cognates information at different levels of machine translation and the lessons that can be drawn from this. Additionally, multilingual neural machine translation of closely related languages is given a particular focus in this survey. This article ends with a discussion of the way forward in machine translation with orthographic information, focusing on multilingual settings and bilingual lexicon induction. 
\end{abstract}
\section{Introduction}
Natural Language Processing (NLP) plays a significant role in keeping languages alive and the development of languages in the digital device era \citewpar{Karakanta2018}. One of the sub-parts of NLP is Machine Translation (MT). MT has been the most promising application of Artificial Intelligence (AI) since the invention of computers, which has has been shown to increase access to information by the native language of the speakers in many cases. One of the such critical case is the spread of vital information during a crisis or emergency \citewpar{lewis-etal-2011-crisis,neubig-hu-2018-rapid}.  Recently, translation accuracy has increased, and commercial systems have gained popularity. These systems have been developed for hundreds of languages, and hundreds of millions of people gained access.  However, some of the less common languages do not enjoy this availability of resources. These under-resourced languages lack essential linguistics resources, e.g. corpora, POS taggers, grammars.  This is more pertinent for MT since most common systems require large amounts of high-quality parallel resources or linguistics experts to make a vast set of rules. This survey studies how to take advantage of the orthographic information and closely related languages to improve the translation quality of under-resourced languages.

The most common MT systems are based on Rule-Based Machine Translation (RBMT) or Corpus-Based Machine Translation (CBMT). RBMT systems \citewpar{kaji-1988-efficient, charoenpornsawat-etal-2002-improving,abercrombie-2016-rule,susanto-etal-2012-rule,centelles-costa-jussa-2014-chinese, allauzen-etal-2014-pushdown,hurskainen-tiedemann-2017-rule} are based on linguistic knowledge which are encoded by experts.  On the other hand, CBMT \citewpar{dauphin-lux-1996-corpus,carl-2000-model} depends on a large number of aligned sentences such as Statistical Machine Translation (SMT) \citewpar{kondrak-etal-2003-cognates,setiawan-etal-2005-phrase,Koehn:2005,koehn2007moses, green-etal-2014-empirical,junczys-dowmunt-grundkiewicz-2016-phrase} and Neural Machine Translation (NMT) \citewpar{Sutskever:2014:SSL:2969033.2969173, cho2014learning, Bahdanau2016, zhang-etal-2017-incorporating}. Unlike RBMT systems, which require expertise of linguists to write down the rules for the language, CBMT-based systems rely on examples in the form of sentence aligned parallel corpora. CBMT systems such as SMT and NMT have alleviated the burden of writing down the rules which are not feasible for all languages since human languages are more dynamic in nature. 

However, CBMT systems suffer from the lack of parallel corpora for under-resourced languages to train machine translation systems. A number of the methods have been proposed to address the non-availability of parallel corpora for under-resourced languages, such as pivot-based approaches \citewpar{wu-wang-2007-pivot,wu-wang-2009-revisiting,kim-etal-2019-pivot}, zero-shot translation \citewpar{johnson-etal-2017-googles, tan-etal-2019-multilingual,gu-etal-2019-improved, pham-etal-2019-improving, currey-heafield-2019-zero} and unsupervised methods \citewpar{artetxe-etal-2019-effective,pourdamghani-etal-2019-translating, artetxe-etal-2019-bilingual}, which are described in detail in following sections. A large array of techniques have been applied to overcome the data sparsity problem in MT, and virtually all of them seem to be based on the field of transfer learning from high-resource languages in recent years. Other techniques are based on lexical and semantic similarities of closely related languages which are more relevant to our survey on orthographic information in machine translation. 

The main goal of this survey is to shed light on how orthographic information is utilised in the MT system development and how orthography helps to overcome the data sparsity problem for under-resourced languages. More particularly, it tries to explain the nature of interactions with orthography with different types of machine translation.  For the sake of simplicity, the analysis presented here in this article is restricted to those languages which have some form of internet resources. The survey is organised as follows: Section \ref{background} explains the background information to follow this article. We present orthographic information in Section \ref{orthographic}. Section \ref{RBMT} describes the challenges of automatically using orthographic information in RBMT outputs.  Section \ref{SMT} presents an analysis of orthographic information in SMT systems.  Section \ref{NMT} presents an analysis of orthographic information in NMT systems. This survey ends with a discussion of the future directions towards utilising the orthographic information.

\section{Background} \label{background}
In this section, we explain the necessary background information to follow the paper, different types of MT systems and the orthographic information available for MT. 
\subsection{Under-resourced Languages}
Worldwide, there are around 7,000 languages \citewpar{P10-1010, L14-1618}. However, most of the machine-readable data and natural language applications are available for very few popular languages, such as Chinese, English, French, or German. For other languages, resources are scarcely available and, for some languages, not at all. Some examples of these languages do not even have a writing system \citewpar{W06-0605,krauwer2003basic,alegria2011strategies}, or are not encoded in major schemes such as Unicode. Due to the unavailability of digital resources, many of these languages may go extinct. With each language that is lost, we lose connection with the culture of the people and characteristics of the languages.  

\citewopar{alegria2011strategies} proposed six levels language typology to develop language technologies that could be useful for several hundred languages. This classifies the world's languages based on the availability of Internet resources for each language. According to the study, the term resource-poor or under-resourced is relative and also depends on the year. The first level is the most resourced languages; the second level is languages in the top 10 languages used on the web. The third level is languages which have some form of resources in NLP. The fourth level considers languages which have any lexical resources. Languages that have a writing system but not in digital form are in the fifth level. The last level is significant, including oral languages which do not have a writing system of its own. We follow this approach to define the term under-resourced languages in terms of machine translation by taking the languages in the third and fourth level. Languages that lack extensive parallel corpora are known as under-resourced or low-resourced languages \citewpar{JIMERSON18.749}. 

Languages that seeks to survive in modern society need NLP, which requires a vast amount of data and linguistic knowledge to create new language technology tools for languages. Mainly it is a big challenge to develop MT systems for these languages due to the scarcity of data, specifically sentence aligned data (parallel corpora) in large amounts to train MT systems.  For example, Irish, Scottish Gaelic, Manx or Tamil, Telugu, and Kannada belonging to the Goidelic and the Dravidian languages, respectively are considered as under-resourced language due to scarcely available machine -readable resources \citewopar{alegria2011strategies}. 

\subsection{Orthographic Information} \label{orthographic}
Humans are endowed with a language faculty that is determined by biological and genetical development. However, this is not true of the written form of the language, which is the visual representation of the natural and genetically determined spoken form. With the development of orthography, humans have not only overcome limitations with human short term memory, and brain storage capacity, but also this development allows communication through space and time \citewpar{fromkin2018introduction}. Orthography is a linguistic factor of mutual intelligibility which may facilitate or impede inter-comprehension \citewpar{fischer2016orthographic}.

The orthographic information of languages does not only represents the information of the language but also the psychological representation of the world of the users. Chinese orthography is unique in its own in the sense that it uses logo-graphic writing system. In such a system, each Chinese character carries visual patterns along with rich linguistic information. These characters are visualised in square space, which depends on the number of strokes a character has. 
Each character can be decomposed in two parts. \textit{Radicals}, which carries the semantic meaning, whereby the other part tells about the pronunciation. According to Shuo WenJie Zi\footnote{https://en.wikipedia.org/wiki/Shuowen$\_$Jiezi} new Chinese characters consist of 540 radicals but only 214 in modern Chinese \cite{min2004direct}. The problems lie when the decomposition strategy does not comply with some of the characters. On the other hand, other Asian languages such as Korean and Japanese, have two different writing systems. Modern-day Korea uses the Hangul orthography, which is part of the syllabic writing system, and the other is known as Hanja, which uses classical Chinese characters.
Like the history of writing in Korea, Japan to have two writing systems, Kana and Kanji, where Kanji is identified as Classical Chinese characters, and Kana represents sounds where each kana character is recognized as a syllable. As both Korean and Japanese are very different from Chinese and morphologically rich languages adoption of Chinese characters to the languages was rather difficult. These problems also posed great difficulty in the field of translation and transliteration. Irrespective of all the differences and challenges these three Asian languages share common properties which could be significant advantages in MT.

Closely related languages share similar morphological, syntactical, orthographic properties. Orthographic similarity can be seen from two major sources. First one is based on the genetic relationship between languages such as based on language families, Germanic, Slavic, Gaelic and Indo-Aryan languages. The second one is based on the contact though geographical area Indo-Aryan and Dravidian languages in the Indian subcontinent \citewpar{kunchukuttan-etal-2018-leveraging}. Two languages posses orthographic similarity only when these languages have the following properties: overlapping phonemes, mutually compatible orthographic systems and similar grapheme to phoneme mapping.

The widespread and underlying problem for the MT systems is variations in orthographic conventions. The two languages written in two different orthography leads to error in MT outputs. Orthographic information can also be used to improve the machine translation system. In the following subsection, we describe the different orthographic properties related to MT.

\subsubsection{Spelling and Typographical Errors}
Spelling or typographical errors are to be handled very carefully in MT task as even a minor spelling error could generate out of vocabulary error in the training corpus. The source and the target languages highly influenced the methodology used to correct orthographic errors. As these languages vary in use of the same orthographic conventions very differently. 

\subsubsection{True-casing and Capitalization}
The process of restoring case information to badly cased or not cased text is true-casing \citewpar{lita2003truecasing}.  To avoid orthographical errors, it is a popular method to lower-case all words, especially in SMT. This method allows the system to avoid the mismatching of the same words, which seems different due to differences in casing thus keeping all the text in the lower-case is one of the methods to avoid the error. In most MT systems, both a pre-processing and post-processing is carried out. Post-processing of the text involves converting all the lower case to its original case form and generating the proper surface forms.

\subsubsection{Normalization}
The use of the same words with different orthographic spellings such as \textit{colour} and \textit{color} give rise to different errors while building a translation model. In such cases, orthographic normalization is required. There are several other issues which require orthographic normalization, which could be language-specific such as Arabic diacritization, or contextual orthographic normalization. This approach needs some linguistic knowledge and can be adapted easily to other languages as well. Normalization is a  process which is carried out before most of the natural language processing task; similarly, in machine translation, language-specific normalization yields a good result. Some of the examples of text normalization carried out for  SMT system are removal of HTML contents, extraction of tag contents, splitting each line after proper punctuation marks as well as correction of language-specific word forms \citewpar{schlippe2010text}. Normalization reduces sparsity as it eliminates out-of-vocabulary words used in the text \citewpar{leusch-etal-2005-preprocessing}.

\subsubsection{Tokenization and Detokenization}
The process of splitting text into smaller elements is known as tokenization. Tokenization can be done at different levels depending on the source and the target language as well the goal that we want to achieve. It also includes processing of the signs and symbols used in the text such as hyphens, apostrophes, punctuation marks, and numbers to make the text more accessible for further steps in MT. Like normalization, tokenization also helps in reducing language sparsity. 

Detokenization is the process of combining all the token to the correct form before releasing the main output. Tokenization and detokenization are not linked directly to orthographic correction, rather, they are more about morphological linking and correction, especially towards morphological rich languages like Irish and Arabic \citewpar{guzman-etal-2016-machine-translation}. Orthography plays a major role in tokenization and detokenizations as each orthography has different rules on how to tokenize and detokenize.

\subsubsection{Transliteration}
Transliteration is the conversion of the text from one orthography to another without any phonological changes. The best example of transliteration is named entities and generic words \citewpar{kumaran2007generic}. Data collected from social media are highly transliterated and contains errors thus, using these data for building a machine translation system for resource-poor languages causes errors. One of the primary forms that have a high chance of transliteration is cognates. Cognates are words from different languages derived from the same root. The concept cognate in NLP approaches are the words with similar orthography. Therefore, cognates have a high chance of transliteration. Though Machine translation has progressed a lot in recently, the method of dealing with transliteration problem has changed from a language-independent manner to cognates prediction when translating between closely related languages, transliteration of cognates would help to improve the result for under-resourced languages. 

\subsubsection{Code-Mixing}
Code-mixing is a phenomenon which occurs commonly in most multilingual societies where the speaker or writer alternate between more than one languages in a sentence \citewpar{ayeomoni2006code,Ranjan2016ACS,yoder2017code,PARSHAD2016375}. Most of the corpora for under-resourced languages came from the publicly available parallel corpora which were created by voluntary annotators or aligned automatically.  The translation of technical documents such as KDE, GNOME, and Ubuntu translations have code-mixed data since some of the technical terms may not be known to voluntary annotators for translation. Code-mixing in the OpenSubtitles corpus is due to bilingual and historical reasons of native speakers \citewpar{chanda2016columbia,PARSHAD2016375}.  Different combinations of languages may occur while code-mixing, for example, German-Italian and French-Italian in Switzerland, Hindi-Telugu in state of Telangana, India, Taiwanese-Mandarin Chinese in Taiwan \citewpar{chan-etal-2009-automatic}. As a result of code-mixing of the script are also possible from a voluntary annotated corpus. This poses another challenge for MT
\section{Orthographic Information in RBMT} \label{RBMT}
Rule-Based Machine Translation (RBMT) was one of the first approaches to tackle translation from the input of the source text to target text without human assistance by means of collections of dictionaries, collections of linguistics rules and special programs based on these dictionaries and rules. It also depends on rules and linguistic resources, such as bilingual dictionaries, morphological analysers, and part-of-speech taggers. The rules dictate the syntactic knowledge while the linguistics resources deal with morphological, syntactic, and semantic information. Both of them are grounded in linguistic knowledge and generated by linguistics experts \citewpar{slocum1985evaluation, charoenpornsawat-etal-2002-improving, Lagarda:2009:SPR:1620853.1620913,  susanto-etal-2012-rule}.  The strength of RBMT is that analysis can be done at both syntax and semantics level.  However, it requires a linguistic expert to write down all the rules that cover languages.

Open-source shallow-transfer MT engine for the Romance languages of Spain such as Spanish, Catalan and Galician developed by  \citewopar{rbmt_spanish}. They were regeneration of existing non-open-source engines based on linguistic data.  The post-generator in the system performs the orthographical operation such as contraction and apostrophes to reduce the orthographical errors. The dictionaries were used for string transformation operations to the target language surface forms. Similarly, the translation between Spanish-Portugues used a post-generation module to performs orthographical transformations to improve the translation quality \citewpar{garrido2004shallow,forcada2011apertium}. 

Manually constructed list of orthographic transformation rules assist in detecting cognates by string matching \citewpar{xu-etal-2015-detecting}.
Irish, Scottish and Gaelic belong to the Goidelic language family and share similar orthography and cognates. \citewopar{scannell2006machine} developed ga2gd software which translates from Irish to Scottish Gaelic. In the context-sensitive syntactic rewriting submodule, the authors implemented transfer rules based on orthography, which are stored in a plain text. Then each rule is transformed into a finite-state recogniser for the input stream. This work also uses simple rule-based orthographic changes to find cognates by orthography.

A Czech to Polish translation system also followed the shallow-transfer method at the lexical stage. A set of collective transformation rules were used on a source language list to produce a target language list of cognates \citewpar{ruth2011shallow}. Another shallow-transfer MT system used frequent orthographic changes from Swedish to Danish to identify cognates and transfer rules are based on orthography \citewpar{tyers2009shallow}. A Turkmen to Turkish MT system \citewpar{tantuug2007mt,tantuug2018machine} uses the finite-state transformer to identify the cognate even thought the orthography is different for these languages.

\section{Orthographic Information in SMT} \label{SMT}
Statistical Machine Translation (SMT) \citewpar{brown-etal-1993-mathematics,Koehn:2005, koehn2007moses, koehn2009statistical, waite-byrne-2015-geometry} is one of the CBMT based systems. SMT systems assume that we have a set of example translations($S^{(k)}$, $T^{(k)}$) for $k=1\ldots.n$, where $S^{(k)}$ is the $k^{th}$ source sentence, $T^{(k)}$ is the $k^{th}$ target sentence which is the translation of $S^{(k)}$ in the corpus. SMT systems try to maximize the conditional probability $p(t|s)$ of target sentence $t$ given a source sentence $s$ by maximizing separately a language model $p(t)$ and the inverse translation model $p(s|t)$. A language model assigns a probability $p(t)$ for any sentence t and translation model assigns a conditional probability $p(s|t)$ to source / target pair of sentence \citewpar{wang-waibel-1997-decoding}. By Bayes rule
\begin{equation}
p(t|s) \propto  p(t)p(s|t)
\end{equation}
This decomposition into a translation and a language model improves the fluency of generated texts by making full use of available corpora. The language model is not only meant to ensure a fluent output, but also supports difficult decisions about word order and word translation \citewpar{koehn2009statistical}.

The two core methodologies used in the development of machine translation systems - RBMT and SMT - come with their own shares of advantages and disadvantages. In the initial stages, RBMTs were the first commercial systems to be developed. These systems were based on linguistic rules and have proved to be more feasible for resource-poor languages with little or no data. It is also relatively simpler to carry out error analysis and work on improving the results. Moreover, these systems require very little computational resources.

On the contrary, SMT systems need a large amount of data, but no linguistic theories, especially with morphologically rich languages such as Irish, Persian, and Tamil SMT suffer from out-of-vocabulary problems very frequently due to orthographic inconsistencies. To evade the problem, orthographic normalization was proposed to improve the quality of SMT by sparsity reduction \citewpar{kholynizar2012}. SMT learns from data and requires less human effort in terms of creating linguistics rules. SMT systems, unlike RBMT system, does not cause disambiguation problems. Even though SMT has lots of advantages over rule-based, it also has some disadvantages. Its is very difficult to conduct error analysis with SMT and data sparsity another disadvantage faced by SMT \citewpar{costa2012study}.

\subsection{Spelling and Typographical Errors }
The impact of spelling and typographical errors in SMT has been studied extensively 
\citewpar{4637895,bertoldi-etal-2010-statistical,formiga-fonollosa-2012-dealing}. Dealing with random, non-word error or real-word error can be done in many ways; one such method is the use of a character-level translator, which provides various spelling alternative. Typographical errors such as substitution, insertion, deletion, transposition, run-on, and split can be addressed with edit-distance under a noisy channel model paradigm \citewpar{brill-moore-2000-improved,toutanova-moore-2002-pronunciation}. Error recovery was performed to correct spelling alternative of input before the translation process.

\subsection{True-casing and Capitalization, Tokenization and Detokenization }
Most SMT systems accept pre-processed inputs, where the pre-processing consists of tokenising, true-casing, and normalising punctuation. Moses \citewpar{koehn2007moses} is a toolkit for SMT, which has pre-processing tools for most languages based on hand-crafted rules. Improvement has been achieved for recasing and tokenization processes \citewpar{nakov-2008-improving}. For a language which does not use Roman characters, linguistically-motivated tokenization has shown to improve the results on SMT \citewpar{oudah-etal-2019-impact}.  Byte Pair Encoding (BPE) avoids out-of-vocabulary issues by representing more frequent sub-word as atomic units \citewopar{sennrich-etal-2016-improving}. A joint BPE model based on the lexical similarity between Czech and Polish identified cognate vocabulary of sub-words. This is based on the orthographic correspondences from which words in both languages can be composed \citewpar{chen-avgustinova-2019-machine}.  

\subsection{Normalization}
Under-resourced languages utilise corpora from the user-generated text, media text or voluntary annotators. However, SMT suffers from customisation problems as tremendous effort is required to adapt to the style of the text. A solution to this is text normalization, that is normalising the corpora before passing it to SMT
\citewpar{formiga-fonollosa-2012-dealing} which has been shown to improve the results. The orthographies of the Irish and Scottish Gaelic languages were quite similar due to a shared literary tradition. Nevertheless, after the spelling reform in Irish, the orthography became different. \citewopar{scannell-2014-statistical} proposed a statistical method to normalise the orthography between Scottish Gaelic and Irish as part of the translation for social media text.  To able to use the current NLP tool to deal with historical text, spelling normalization is essential; that is converting the original spelling to present-day spelling which was studied for historical English text by \citewopar{schneider-etal-2017-comparing} and \citewopar{hamalainen-etal-2018-normalizing}. For dialects translation, spelling normalising is an important step to take advantage of high-resource languages resources \citewpar{honnet-etal-2018-machine,napoles-callison-burch-2017-systematically}

\subsection{Transliteration (Cognate) }
As we know, closely related languages share the same features; the similarities between the language would be of much help to study the cognates of two languages. Several methods have been obtained to manipulate the features of resource-rich languages in order to improve SMT for resource-poor languages. Manipulation of the cognates to obtain transliteration is one of the methods adopted by some of the authors to improve the SMT system for resource-poor languages.

Language similarities and regularities in morphology and spelling variation motivate the use of character-level transliteration models. However, in order to avoid the character mapping differences in various contexts \citewopar{nakov2012combining} transformed the input to a sequence of character n-grams. A sequence character of n-grams increases the vocabulary as well as also make the standard alignment models and their lexical translation parameters more expressive.

For the languages which use same or similar scripts, approximate string matching approaches, like Levenshtein distance \citewpar{Levenshtein_SPD66} are used to find cognate and longest common subsequence ratio (LCSR) \citewpar{melamed-1999-bitext}.  For the languages which use different scripts, transliteration is the first step and follow the above approach. A number of studies have used statistical and deep learning methods along with orthographic information \citewpar{ciobanu-dinu-2014-automatic,mulloni-pekar-2006-automatic} to find the cognates. In reference to the previous section we know that cognates can be used for mutual translation between two languages if they share similar properties, it is essential to know the cognateness between the two languages of a given text. The word "cognateness" means how much two pieces of text are related in terms of cognates. These cognates were useful to improve the alignment when the scoring function of the length-based alignment function is very low then it passes to the second method, a cognate alignment function for getting a proper alignment result \citewpar{simard1993using}.

One of the applications of cognates before applying MT is parallel corpora alignment.  A study of using cognates to align sentences for parallel corpora was done by \citewopar{10.5555/962367.962411}. Character level methods to align sentences \citewpar{church-1993-char} are based on a cognate approach \citewpar{10.5555/962367.962411}.

As early as \citewopar{C88-1010}, researchers have looked into translation between closely-related languages such as from Czech-Russian RUSLAN and Czech-Slovak CESILKO \citewpar{A00-1002} using syntactic rules and lexicons. The closeness of the related languages makes it possible to obtain a good translation by means of more straightforward methods. However, both systems were rule-based approaches and bottlenecks included complexities associated with using a word-for-word dictionary translation approach. Nakov and Ng~\citewpar{D09-1141} proposed a method to use resource-rich closely-related languages to improve the statistical machine translation of under-resourced languages by merging parallel corpora and combining phrase tables. The authors developed a transliteration system trained on automatically-extracted likely cognates for Portuguese into Spanish using systematic spelling variation.  

\citewopar{W14-4210} created an MT system between closely-related languages for the Slavic language family. Language-related issues between Croatian, Serbian and Slovenian are explained by \citewopar{W16-4806}. Serbian is digraphic (uses both Cyrillic and Latin Script), the other two are written using only the Latin script. For the Serbian language transliteration without loss of information is possible from Latin to Cyrillic script because there is a one-to-one correspondence between the characters. 

In 2013 a group of people used a PBSMT approach as the base method to produce cognates. Instead of translating the phrase, they tried to transform a character sequence from one language to another. They have used words instead of sentences and characters instead of words in the transformation process. The combination of the phrase table with transformation probabilities, language model probabilities, selects the best combination of sequence. Thus the process includes the surrounding context and produces cognates \citewpar{beinborn2013cognate}. A joint BPE model based on the lexical similarity between Czech and Polish identifies a cognate vocabulary of sub-words. This is based on the orthographic correspondences from which words in both languages can be composed \citewpar{chen-avgustinova-2019-machine}.  It has been demonstrated that the use of cognates improves the translation quality \citewpar{kondrak-etal-2003-cognates}.

\subsection{Code-Switching}
An SMT system with a code-switched parallel corpus was studied by \citewopar{menacer2019machine} and \citewopar{fadaee-monz-2018-back} for Arabic-English language pair. The authors have manually translated or used back translation method to translate foreign words. The identification of the language of the word is based on the orthography. \citewopar{chakravarthi2018improving} used the same approach for Dravidian languages; they used the improved MT for creating WordNet, showing improvement in the results. For English-Hindi, \citewopar{dhar-etal-2018-enabling} manually translated the code-switched component and shown improvements. Machine translation of social media was studied by \citewopar{rijhwani2016translating} where they tackle the code-mixing for Hindi-English and Spanish-English. The same approach translated the main language of the sentence using Bing Translate API \citewpar{niu-etal-2018-bi}.

Back transliteration from one script to native script in code-mixed data is one of the challenging tasks to be performed. \citewopar{riyadh2019joint} adopted three different methods to back transliterate Romanised Hindi-Bangla code-mixed data to Hindi and Bangla script. They have used Sequitur, a generative joint n-gram transducer, DTLM, a discriminate string transducer and the OpenNMT \footnote{https://opennmt.net/} neural machine translation toolkit. Along with these three approaches, they have leveraged target word lists, character language models, as well as synthetic training data, whenever possible, in order to support transliteration. At last, these transliterations are provided to a sequence prediction module for further processing.

\subsection{Pivot Translation}
Pivot translation is a translation from a source language to the target language through an intermediate language which is called a pivot language. Usually, pivot language translation has large source-pivot and pivot-target parallel corpora \citewpar{cohn-lapata-2007-machine, wu-wang-2009-revisiting}.  There are different levels of pivot translation, the first one is the triangulation method where the corresponding translation probabilities and lexical weights in the source-pivot and pivot-target translation are multiplied. In the second method, the sentences are translated to the pivot language using the source-pivot translation system then pivoted to target language using a pivot-target translation system \citewpar{utiyama-isahara-2007-comparison}. Finally, using the source-target MT system to create more data and adding it back to the source-target model, which is called back-translation \citewpar{sennrich-etal-2016-improving, edunov-etal-2018-understanding}.  Back translation is simple and easy to achieve without modifying the architecture of the machine translation models. Back-translation has been studied in both SMT \citewpar{tiedemann-etal-2016-phrase,ahmadnia-etal-2017-persian,poncelas-etal-2019-combining} and NMT \citewpar{sennrich-etal-2016-improving, edunov-etal-2018-understanding, hoang-etal-2018-iterative, prabhumoye-etal-2018-style, graca-etal-2019-generalizing,kim-etal-2019-pivot}.

The pivot translation method could also be used to improve MT systems for under-resourced languages. One popular way is training SMT systems using source-pivot or pivot-target language pair using sub words where the pivot language is related to source or target or both. The subwords units consisted of orthographic syllable and byte-pair-encoded unit.  The orthographic unit is a linguistically motivated unit which occurs in a sequence of one or more consonants followed by a vowel. Unlike orthographic units, BPE (Byte Pair Encoded Unit) \citewpar{sennrich-etal-2016-improving} is motivated by statistical properties of the text. It represents stable and frequent character sequences in the texts. As orthographic syllable and BPE are variable-length units and the vocabularies used are much smaller than morpheme and word-level model, the problem of data sparsity does not occur but provides an appropriate context for translation between closely related languages \citewpar{kunchukuttan2017utilizing}.

\section{Orthographic Information in NMT} \label{NMT}
Neural Machine Translation is a  sequence-to-sequence approach \citewpar{Sutskever:2014:SSL:2969033.2969173} based on encoder-decoder architectures with attention  \citewpar{Bahdanau2016, saunders-etal-2019-domain} or self attention encoder \citewpar{Vaswani:2017:AYN:3295222.3295349,wang-etal-2019-learning}. Given a source sentence $\mathbf{x}$=${x_1,x_2,x_3,...}$ and target sentence $\mathbf{y}$=${y_1,y_2,y_3,..}$, the training objective for NMT is to maximize the log-likelihood $\mathcal{L}$ with respect to $\theta$:
\begin{equation}
\mathcal{L}_{\theta}=\sum_{(\mathbf{x}, \mathbf{y}) \in \mathrm{C}} \log p(\mathbf{y} | \mathbf{x} ; \theta)
\end{equation}
The decoder produces one target word at a time by computing the probability 
\begin{equation}
p(\mathbf{y} | \mathbf{x} ; \theta)=\prod_{j=1}^{m} p\left(y_{j} | y_{<j}, \mathbf{x} ; \theta\right)
\end{equation}

Where $m$ is the number of words in $\mathbf{y}, y_{j}$ is the current generated word, and $y_{<j}$ are the previously generated words. At inference time, beam search is typically used to find the translation that maximises the above probability. Most of NMT models follows the $Embedding\rightarrow$ $Encoder\rightarrow$ $Attention\rightarrow$ $Decoder$ framework.

The attention mechanism across encoder and decoder  is calculated by $c_t$ as the weighted sum of the source-side context vectors:
\begin{equation}
    c_t=\sum_{i=1}^n \alpha_{t,i} h_i
\end{equation}

\begin{equation}
        \alpha_{t,i}= \frac{\exp{(e_{t,i}})}{\sum_{j=1}^{m}\exp{(e_{t,j}})}
\end{equation}
$\alpha_{t,i}$ is the normalized alignment matrix between each source annotation vector $h_i$ and word $y_t$ to be emitted at a time step $t$. Expected alignment $e_{t,i}$ between each source annotation vector $h_i$ and the target word $y_t$ is computed using the following formula:
\begin{equation}
    e_{t,i}=a(\mathbf{s}_{\mathbf{t}-\mathbf{1}},h_i) 
\end{equation}

\begin{equation}
\mathbf{s}_{\mathbf{t}}=g\left(\mathbf{s}_{\mathbf{t}-\mathbf{1}}, \mathbf{y}_{\mathbf{t}-\mathbf{1}}, \mathbf{c}_{\mathbf{t}}\right)
\end{equation}

where $g$ is an activation decoder function, $s_{j-1}$ is the previous decoder hidden state, $y_{j-1}$ is the embedding of the previous word. The current decoder hidden state $s_{j},$ the previous word embedding and the context vector are fed to a feedforward layer $f$ and a softmax layer computes a score for generating a target word as output:
\[
P\left(y_{j} | y_{<j}, \mathbf{x}\right)=\operatorname{softmax}\left(f\left(\mathbf{s}_{\mathbf{j}}, \mathbf{y}_{j-1}, \mathbf{c}_{\mathbf{j}}\right)\right)
\]

\subsection{Multilingual Neural Machine Translation}
In recent years, NMT has improved translation performance, which has lead to a boom in NMT research. The most popular neural architectures for NMT are based on the encoder-decoder \citewpar{Sutskever:2014:SSL:2969033.2969173,cho-etal-2014-learning,Bahdanau2016} structure and the use of attention or self-attention based mechanism \citewpar{luong-etal-2015-effective,Vaswani:2017:AYN:3295222.3295349}. Multilingual NMT created with or without multiway corpora has been studied for the potential for translation between two languages without any direct parallel corpus. Zero-shot translation is translation using multilingual data to create a translation for languages which have no direct parallel corpora to train independently. Multilingual Neural Machine Translation with only monolingual corpora was studied by
\citewpar{sen-etal-2019-multilingual, wang-etal-2019-compact}. In \citewopar{DBLP:journals/corr/HaNW16} and \citewpar{johnson-etal-2017-googles}, the authors have demonstrated that multilingual NMT improves translation quality. For this, they created a multilingual NMT without changing the architecture by introducing special tokens at the beginning of the source sentence indicating the source language and target language.

Phonetic transcription to Latin script and the International Phonetic Alphabet (IPA) was studied by \citewopar{chakravarthi2018improving} and showed that Latin script outperforms IPA for the Multilingual NMT of Dravidian languages. \citewopar{chakravarthi-et-al:OASIcs:2019:10370} propose to combine multilingual, phonetic transcription and multimodal content with improving the translation quality of under-resourced Dravidian languages. The authors studied how to use the closely-related languages from the Dravidian language family to exploit the similar syntax and semantic structures by phonetic transcription of the corpora into Latin script along with image feature to improve the translation quality \citewpar{chakravarthi2019wordnet}.  They showed that orthographic information improves the translation quality in multilingual NMT\citewpar{chakravarthi-etal-2019-multilingual}.

\subsection{Spelling and Typographical Errors}
Spelling errors are amplified in under-resourced setting due to the potential infinite possible misspelling and leads to a large number of out-of-vocabulary. Additionally, under-resourced morphological rich languages have morphological variation, which causes orthographic errors while using character level MT. A shared task was organised by \citewopar{li-etal-2019-findings}; 
to deal with orthographic variation, grammatical error and informal languages from the noisy social media text.  Data cleaning was used along with suitable corpora to handle spelling errors. \citewopar{belinkov2018synthetic} investigated noise in NMT, focusing on kinds of orthographic errors. Parallel corpora were cleaned before submitting to NMT to reduce the spelling and typographical errors.

NMT with word embedding lookup ignores the orthographic representation of the words such as the presence of stems, prefixes, suffixes and another kind of affixes. To overcome these drawbacks, character-based word embedding was proposed by \citewopar{10.5555/3016100.3016285}. Character-based NMT \citewpar{costa-jussa-fonollosa-2016-character,yang-etal-2016-character,lee-etal-2017-fully,cherry-etal-2018-revisiting} were developed to cover the disadvantages of the languages which do not have explicit word segmentation. This enhances the relationship between the orthography of a word and its meaning in the translation system. For spelling mistake data for under-resourced languages, the quality of word-based translation drops severely, because every non-canonical form of the word cannot be represented. Character-level model overcomes the spelling and typological error without much effort.
\subsection{True-casing and Capitalization, Normalization, Tokenization and Detokenization}
Although NMT can be trained end-to-end translations, many NMT systems are still language-specific and require language-dependent preprocessing, such as used in Statistical Machine Translation, Moses \citewpar{koehn2007moses} a toolkit for SMT which has preprocessing tools for most languages which based on hand-crafted rules. In fact, these are mainly available for European languages. For Asian languages which do not use space between words, a segmenter is required for each language independently before feeding into NMT to indicate a word segment. This becomes a problem when we train Multilingual NMT \citewpar{johnson-etal-2017-googles}.

A solution for the open vocabulary problems in NMT is to break up the rare words into subword units \citewpar{chitnis-denero-2015-variable, ding-etal-2019-call}
which has been shown to deal with multiple script languages ambiguities \citewpar{6289079,wu2016google}. A simple and language-independent tokenizer was introduced for NMT and Multilingual NMT by \citewopar{kudo-richardson-2018-sentencepiece}; it is based on two subword segmentation algorithms, byte-pair encoding (BPE) \citewpar{sennrich-etal-2016-improving} and a unigram language model  \citewpar{kudo-2018-subword}. This system also normalise semantically equivalent Unicode character into canonical forms. Subword segmentation and true-casing model will be rebuilt whenever the training data changes. The preprocessing tools introduced by OpenNMT normalises characters and separates punctuation from words, and it can be used for any languages and any orthography \citewpar{2017opennmt}. 

Character-level NMT systems work at the character level to grasp orthographic similarity between the languages. They were developed to overcome the issue of limited parallel corpora and resolve the out-of-vocabulary problem for the under-resourced languages. For Hindi-Bhojpuri, where Bhojpur is closely related to Hindi, Bhojpuri is considered as an under-resourced language, and it has an overlap of word with high-resource language Hindi due to the adoption of works from a common properties of language
\citewpar{jha2019learning}. To solve the out-of-vocabulary problem the transduction of Hindi word to Bhojpuri words was adapted from NMT models by training on Hindi-Bhojpuri cognate pairs. It was a two-level system: first, the Hindi-Bhojpuri system was developed to translate the sentence; then the out-of-vocabulary words were transduced.

\subsection{Transliteration (Cognate)}
Transliteration emerged to deal with proper nouns and technical terms that are translated with preserved pronunciation. Transliteration can also be used to improve machine translation between closely related languages, which uses different scripts since closely related languages language have orthographic and phonological similarities between them.  

Machine Translation often occurs between closely related languages or through a pivot language (like English) \citewpar{bhattacharyya-etal-2016-statistical}. Translation between closely related languages or dialects is either a simple transliteration from one language to another language or a post-processing step.   Transliterating cognates has been shown to improve MT results since closely related languages share linguistic features. To translate from English to Finnish and Estonian, where the words have similar orthography \citewopar{gronroos-etal-2018-cognate} used Cognate Morfessor, a multilingual variant of Morfessor which learns to model cognates pairs based on the unweighted Levenshtein distance \citewpar{Levenshtein_SPD66}. The ideas are to improve the consistency of morphological segmentation of words that have similar orthography, which shows improvement in the translation quality for the resource-poor Estonian language. 

\citewopar{D09-1111} use transliteration as a method to handle out-of-vocabulary (OOV) problems. To remove the script barrier, \citewopar{DBLP:conf/coling/BhatBJS16} created machine transliteration models for the common orthographic representation of Hindi and Urdu text. The authors have transliterated text in both directions between Devanagari script (used to write the Hindi language) and Perso-Arabic script (used to write the Urdu language). The authors have demonstrated that a dependency parser trained on augmented resources performs better than individual resources. The authors have shown that there was a significant improvement in BLEU (Bilingual Evaluation Understudy) \citewpar{P02-1040} score and have shown that the problem of data sparsity is reduced.

Recent work by \citewopar{Q18-1022} has explored orthographic similarity for transliteration. In their work, they have used related languages which share similar writing systems and phonetic properties such as Indo-Aryan languages. They have shown that multilingual transliteration leveraging similar orthography outperforms bilingual transliteration in different scenarios.  Phonetic transcription is a method for writing a language in the other scripts keeping the phonemic units intact. It is extensively used in speech processing research, text-to-speech, and speech database construction — phonetic transcription to common script has shown to improve the results of machine translation \citewpar{chakravarthi2018improving}. The authors focus on the multilingual translation of languages which uses different scripts and studies the effect of different orthographies to common script with multilingual NMT.  
Multiway NMT system was created for Czech and Polish with Czech IPA transcription and Polish transcription to a 3-way parallel text together to take advantage of the phonology of the closely related languages \citewpar{chen-avgustinova-2019-machine}. Orthographic correspondence rules were used as a replacement list for translation between closely related Czech-Polish with added back-translated corpus \citewpar{chen-avgustinova-2019-machine}. Dialect translation was studied by \citewopar{baniata2018neural}. To translate Arabic dialects to modern standard Arabic, they used multitask learning which shares one decoder for standard Arabic, while every source has a separate encoder. This is due to the non-standard orthography in the Arabic dialects. The experiments showed that for the under-resourced Arabic dialects, it improved the results. 

Machine Translation of named entities is a significant issue due to linguistic and algorithmic challenges found in between languages. The quality of MT of named entities, including the technical terms, was improved with the help of developing lexicons using orthographic information. The lexicon integration to NMT was studied for the Japanese and Chinese MT \citewpar{halpern-2018-large}. They deal with the orthographic variation of named entities of Japanese using large scale lexicons.  For English-to-Japanese, English-to-Bulgarian, and English-to-Romanian \citewopar{ugawa-etal-2018-neural} proposed a model that encodes the input word based on its NE tag at each time step. This helps to improve the BLEU score for machine translation results.

\subsection{Code-Switching}
A significant part of corpora for under-resourced languages comes from movie subtitles and technical documents, which makes it even more prone to code-mixing. Most of these corpora are movie speeches \citewpar{birch-etal-2019-global} transcribed to text, and they differ from that in other written genres: the vocabulary is informal, non-linguistics sounds like \textit{ah},  and mixes of scripts in case of English and native languages \citewpar{TIEDEMANN08.484,chakravarthi2016,chakravarthi-code-mix-survey,chakravarthi-etal-2020-senti-tamil,chakravarthi-etal-2020-senti-malayalam,chakravarthi-code-mix-ruba-ne}. 
Data augmentation \citep{fadaee-etal-2017-data,li-specia-2019-improving} and changing the foreign to native words using dictionaries or other methods have been studied. Removing the code-mixing word from the corpus on both sides was studied by \citewopar{chakravarthi2018improving,chakravarthi2019wordnet} for English-Dravidian languages. \citewopar{song-etal-2019-code} studied the data augmentation method, making code-switched training data by replacing source phrases with their target translation. Character-based NMT \citewpar{costa-jussa-fonollosa-2016-character,yang-etal-2016-character,lee-etal-2017-fully} can naturally handle intra-sentence codeswitching as a result of the many-to-one translation task.

\section{Orthographic Information in Unsupervised Machine Translation}
Building parallel corpora for the under-resourced languages is time-consuming and expensive. As a result parallel corpora for the under-resourced languages are limited or unavailable for some of the languages. With limited parallel corpora, supervised SMT and NMT cannot achieve the desired quality translations. However, monolingual corpora can be collected from various sources on the Internet, and are much easier to obtain than parallel corpora. Recent research has created a machine translation system using only monolingual corpora \citewpar{koehnandkhight2000, ravi-knight-2011-deciphering, dou-etal-2014-beyond} by the unsupervised method to remove the dependency of sentence aligned parallel corpora.  These systems are based on both SMT \citewpar{klementiev-etal-2012-toward,artetxe-etal-2018-unsupervised} and NMT \citewpar{artetxe2018iclr}. One such task is bilingual lexicon induction. 

Bilingual lexicon induction is a task of creating word translation from monolingual corpora in two languages \citewpar{turcato-1998-automatically,rosner-sultana-2014-automatic}. One way to induce the bilingual lexicon induction is using orthographic similarity. Based on the assumptions that words that are spelled similarly are sometimes good translation and maybe cognates as they have similar orthography due to historical reasons. A generative model for inducing a bilingual lexicon from monolingual corpora by exploiting orthographic and contextual similarities of words in two different languages was proposed by  \citewopar{haghighi-etal-2008-learning}. Many methods, based on edit-distance and orthographic similarity are proposed for using linguist feature for word alignments supervised and unsupervised methods \citewpar{dyer-etal-2011-unsupervised,berg-kirkpatrick-etal-2010-painless,hauer-etal-2017-bootstrapping}.  \citewopar{riley-gildea-2018-orthographic} proposed method to utilise the orthographic information in word-embedding based bilingual lexicon induction. The authors used the two languages' alphabets to extend the word embedding and modifying the similarity score functions of previous word-embedding methods to include the orthographic similarity measure. Bilingual lexicons are shown to improve machine translation in both RBMT \citewpar{turcato-1998-automatically} and CBMT \citewpar{chu-etal-2014-improving,dou-knight-2013-dependency,dou-etal-2014-beyond}.

In work by \citewopar{W17-2504}, the authors translated lexicon induction for a heavily code-switched text of historically unwritten colloquial words via loanwords using expert knowledge with language information.  Their method is to take word pronunciation (IPA) from a donor language and convert them into the borrowing language. This shows improvements in BLEU score for induction of Moroccan Darija-English translation lexicon bridging via French loan words.  

\section{Conclusion} \label{con}
In this work, we presented a review of the current state-of-the-art in machine translation utilising orthographic information, covering rule-based machine translation, statistical machine translation, neural machine translation and unsupervised machine translation. As a part of this survey, we introduced different machine translations methods and have shown how orthography played a role in machine translation results. These methods to utilise the orthographic information have already let to a significant improvement in machine translation results. 

From our comprehensive survey, we can see that orthographic information improves translation quality in all types of machine translation from rule-based to completely unsupervised systems like bilingual lexicon induction.  For the rule-based machine translation, translation between the closely related language is simplified to transliteration due to the cognates. Statistical machine translation deals with data sparsity problem by using orthographic information. Since statistical machine translation has been studied a long time, most of the orthographic properties are studies for different types of languages. Even the recent neural machine translation and other methods still use preprocessing tools such as true-casers, tokenizers, and detokenizers that are developed for statistical machine translation. Recent neural machine translation is completely end-to-end, however, it suffers from data sparsity when dealing with morphologically rich languages or under-resourced languages. These issues are dealt by utilising orthographic information in neural machine translation. One such method which improves the translation is a transliteration of cognates. Code-switching is another issue with under-resourced languages due to the data collected from voluntary annotator, web crawling or other such methods. However, dealing with code-switching based on orthography or using character-based neural machine translation has been shown to improve the results significantly. 

From this, we conclude that orthographic information is much utilised while translating between closely related languages or using multilingual neural machine translation with closely related languages. While exciting advances have been made in machine translation in recent years, there is still an exciting direction for exploration from leveraging linguistic information to it, such as orthographic information. One such area is unsupervised machine translation or bilingual lexicon induction. Recent works show that word vector, along with orthographic information, performs better for aligning the bilingual lexicons in completely unsupervised or semi-supervised approaches. We believe that our survey will help to catalogue future research papers and better understand the orthographic information to improve machine translation results.

\section*{Acknowledgments}
This publication has emanated from research supported in part by a research grant from Science Foundation Ireland (SFI) under Grant Number SFI/12/RC/2289 (Insight), SFI/12/RC/2289$\_$P2 (Insight$\_$2), \& SFI/18/CRT/6223 (CRT-Centre for Research Training in Artficial Intelligence) co-funded by the European Regional Development Fund as well as by the EU H2020 programme under grant  agreements 731015 (ELEXIS-European Lexical Infrastructure), 825182 (Prêt-à-LLOD), and Irish Research Council grant IRCLA/2017/129 (CARDAMOM-Comparative Deep Models of Language for Minority and Historical Languages). 
\bibliographystyle{spbasic}
\bibliography{main}

\end{document}